\documentclass{article} 
\usepackage{iclr2019_conference,times} 

\usepackage{hyperref}
\usepackage{url}
\usepackage{graphicx}
\usepackage{amsmath}
\usepackage{amssymb}
\usepackage{algorithm,algorithmicx,algpseudocode}
\usepackage{amssymb}
\usepackage{subcaption}

\usepackage{booktabs}
\usepackage{color}
\usepackage{multicol}
\usepackage{arydshln}

\title{Adversarial Information Factorization}

\author{Antonia Creswell \\
Imperial College London\\
\texttt{ac2211@ic.ac.uk} \\
\And
Yumnah Mohamied\\
Imperial College London\\
\texttt{ym1008@ic.ac.uk} \\
\And
Biswa Sengupta\\
AXA-XL Centre of Excellence for Artificial Intelligence\\
\texttt{biswa.sengupta@axaxl.com} 
\AND
Anil A Bharath \\
Imperial College London\\
\texttt{aab01@ic.ac.uk} \\
}

\begin{document}

\maketitle

\begin{abstract}
We propose a novel generative model architecture designed to learn representations for images that factor out a single attribute from the rest of the representation. A single object may have many attributes which when altered do not change the identity of the object itself. Consider the human face; the identity of a particular person is independent of whether or not they happen to be wearing glasses. The \textit{attribute} of wearing glasses can be changed without changing the \textit{identity} of the person. However, the ability to manipulate and alter image attributes without altering the object identity is not a trivial task. Here, we are interested in learning a representation of the image that separates the identity of an object (such as a human face) from an attribute (such as `wearing glasses'). We demonstrate the success of our factorization approach by using the learned representation to synthesize the same face with and without a chosen attribute. We refer to this specific synthesis process as image attribute manipulation. We further demonstrate that our model achieves competitive scores, with state of the art, on a facial attribute classification task.

\end{abstract}

\section{Introduction}

Latent space generative models, such as generative adversarial networks (GANs) \citep{Goodfellow2014,radford2015unsupervised} and variational autoencoders (VAEs) \citep{kingma2013auto}, learn a mapping from a latent encoding space to a data space, for example, the space of natural images. It has been shown that the latent space learned by these models is often organized in a near-linear fashion \citep{radford2015unsupervised, kingma2013auto}, whereby neighbouring points in latent space map to similar images in data space. Certain ``directions'' in latent space correspond to changes in the intensity of certain attributes. In the context of faces, for example, directions in latent space would correspond to the extent to which someone is smiling. This may be useful for image synthesis where one can use the latent space to develop new design concepts \citep{dosovitskiy2017learning,zhu2016generative}, edit an existing image \citep{zhu2016generative} or synthesize avatars \citep{wolf2017unsupervised, taigman2016unsupervised}. This is because semantically meaningful changes may be made to images by manipulating the latent space \citep{radford2015unsupervised,zhu2016generative,larsen2015autoencoding}.

One avenue of research for latent space generative models has been class conditional image synthesis \citep{chen2016infogan, odena2016conditional, mirza2014conditional}, where an image of a particular object category is synthesized. Often, object categories may be sub-divided into fine-grain sub-categories. For example, the category ``dog'' may be split into further sub-categories of different dog breeds. Work by \cite{bao2017cvae} propose latent space generative models for synthesizing images from fine-grained categories, in particular for synthesizing different celebrities' faces conditional on the identity of the celebrity. 

Rather than considering fine-grain categories, we propose to take steps towards solving the different, but related problem of image attribute manipulation. To solve this problem we want to be able to synthesize images and only change one element or attribute of its content. For example, if we are synthesizing faces we would like to edit whether or not a person is smiling. This is a different problem to fine-grain synthesis; we want to be able to synthesize two faces that are similar, with only a single chosen attribute changed, rather than synthesizing two different faces. The need to synthesis two faces that are similar makes the problem of image attribute manipulation more difficult than the fine-grain image synthesis problem; we need to learn a latent space representation that separates an object category from its attributes.

In this paper, we propose a new model that learns a factored representation for faces, separating attribute information from the rest of the facial representation. We apply our model to the CelebA \citep{liu2015deep} dataset of faces and control several facial attributes. 

Our contributions are as follows:

\begin{enumerate}
\item Our \textbf{core contribution} is the novel cost function for training a VAE encoder to learn a latent representation which factorizes binary facial attribute information from a continuous identity representation (Section \ref{sec:IS}).
\item We provide an \textbf{extensive quantitative} analysis of the contributions of each of the many loss components in our model (Section \ref{sec:quant_results}).
\item We obtain classification scores that are competitive with state of the art \citep{zhuang2018multi} using the classifier that is already incorporated into the encoder of the VAE (Section \ref{sec:class}).
\item We provide qualitative results demonstrating that our latent variable, generative model may be used to successfully edit the `Smiling' attribute in more than $90\%$ of the test cases (Section \ref{sec:qual_results}).
\item We discuss and clarify the distinction between conditional image synthesis and image attribute editing (Section \ref{sec:related_work}).
\item We present code to reproduce experiments shown in this paper: (provided after review).
\end{enumerate}

\section{Latent space generative models}
\label{sec:gen-models}

Latent space generative models come in various forms. Two state-of-art generative models are Variational Autoencoders (VAE) \citep{kingma2013auto} and Generative Adversarial Networks (GAN). Both models allow synthesis of novel data samples from latent encodings, and are explained below in more detail.

\subsection{Variational Autoencoder (VAE)}
Variational autoencoders \citep{kingma2013auto} consist of an encoder $q_\phi(z|x)$ and decoder $p_\theta(x|z)$; oftentimes these can be instantiated as neural networks, $E_\phi(\cdot)$ and $D_\theta(\cdot)$ respectively, with learnable parameters, $\phi$ and $\theta$. A VAE is trained to maximize the evidence lower bound (ELBO) on $\log p(x)$, where $p(x)$ is the data-generating distribution. The ELBO is given by:

\begin{equation} \label{eqn:VAE}
\log p(x) \geq \mathbb{E}_{q_\phi(z|x)} \log p_\theta(x|z) -  KL[q_\phi(z|x) || p(z)]
\end{equation}
where $p(z)$ is a chosen prior distribution such as $p(z) = \mathcal{N}(\mathbf{0},I)$. The encoder predicts, $\mu_\phi(x)$ and $\sigma_\phi(x)$ for a given input $x$ and a latent sample, $\hat{z}$, is drawn from $q_\phi(z|x)$ as follows: $\epsilon \sim \mathcal{N}(\mathbf{0},I)$ then $z = \mu_\phi(x) + \sigma_\phi(x) \odot \epsilon$. By choosing a multivariate Gaussian prior, the $KL$-divergence may be calculated analytically \citep{kingma2013auto}. The first term in the loss function is typically approximated by calculating the reconstruction error between many samples of $x$ and $\hat{x} = D_\theta(E_\phi(x))$.

New data samples, which are not present in the training data, are synthesised by first drawing latent samples from the prior, $z \sim p(z)$, and then drawing data samples from $p_\theta(x|z)$. This is equivalent to passing the $z$ samples through the decoder, $D_\theta(z)$. 

VAEs offer both a generative model, $p_\theta(x|z)$, and an encoding model, $q_\phi(z|x)$, which are useful as starting points for image editing in the latent space. However, samples drawn from a VAE are often blurred \citep{radford2015unsupervised}.

\subsection{Generative Adversarial Networks (GAN)}
An alternative generative model, which may be used to synthesize much sharper images, is the Generative Adversarial Network (GAN) \citep{Goodfellow2014, radford2015unsupervised}. GANs consist of two models, a generator, $G_\theta(\cdot)$, and a discriminator, $C_\chi(\cdot)$, both of which may be implemented using convolutional neural networks \citep{radford2015unsupervised,denton2015deep}. GAN training involves these two networks engaging in a mini-max game. The discriminator, $C_\chi$, is trained to classify samples from the generator, $G_\theta$, as being `fake' and to classify samples from the data-generating distribution, $p(x)$, as being `real'. The generator is trained to synthesize samples that confuse the discriminator; that is, to synthesize samples that the discriminator cannot distinguish from the `real' samples. The objective function is given by:

\begin{equation} \label{eqn:aux}
\min_\chi \max_\theta \mathbb{E}_{p(x)} [ \log(C_\chi(x)] + \mathbb{E}_{p_g(x)} [\log (1 - C_\chi(x))]
\end{equation}
where $p_g(x)$ is the distribution of synthesized samples, sampled by: $z \sim p(z)$, then $x = G_\theta(z)$, where $p(z)$ is a chosen prior distribution such as a multivariate Gaussian.

\subsection{Best of both GAN and VAE}
\label{sec:best_of_both}
The vanilla GAN model does not provide a simple way to map data samples to latent space. Although there are several variants on the GAN that do involve learning an encoder type model \citep{dumoulin2016adversarially,Donahue2016,li2017towards}, only the approach presented by \cite{li2017towards} allows data samples to be faithfully reconstructed. The approach presented by \cite{li2017towards} requires adversarial training to be applied to several high dimensional distributions. Training adversarial networks on high dimensional data samples remains challenging \citep{arjovsky2017towards} despite several proposed improvements \citep{salimans2016improved,arjovsky2017wasserstein}. For this reason, rather than adding a decoder to a GAN, we consider an alternative latent generative model that combines a VAE with a GAN. In this arrangement, the VAE may be used to learn an encoding and decoding process, and a discriminator may be placed after the decoder to ensure higher quality of the data samples outputted from the decoder. Indeed, there have been several suggestions on how to combine VAEs and GANs \citep{bao2017cvae, larsen2015autoencoding, mescheder2017adversarial} each with a different structure and set of loss functions, however, none are designed specifically for attribute editing.

The content of image samples synthesized from a vanilla VAE or GAN depends on the latent variable $z$, which is drawn from a specified random distribution, $p(z)$. For a well-trained model, synthesised samples will resemble samples in the training data. If the training data consists of images from multiple categories, synthesized samples may come from any, or possibly a combination, of those categories. For a vanilla VAE, it is not possible to choose to synthesize samples from a particular category. However, conditional VAEs (and GANs) \citep{chen2016infogan,odena2016conditional,mirza2014conditional} provide a solution to this problem as they allow synthesis of class-specific data samples.

\subsection{Conditional VAEs}

Autoencoders may be augmented in many different ways to achieve category-conditional image synthesis \citep{bao2017cvae}. It is common to append a one-hot label vector, $y$, to inputs of the encoder and decoder \citep{sohn2015learning}. However, for small label vectors, relative to the size of the inputs to the encoder and the decoder model, it is possible for the label information, $y$, to be ignored\footnote{The label information in  $y$ is less likely to be ignored when $y$ has relatively high dimensions compared to $z$ \citep{yan2016attribute2image}.}. A more interesting approach, for conditional (non-variational and semi-supervised) autoencoders is presented by \cite{makhzani2015adversarial}, where the encoder outputs both a latent vector, $\hat{z}$, and an attribute vector, $\hat{y}$. The encoder is updated to minimize a classification loss between the true label, $y$, and $\hat{y}$. We incorporate a similar architecture into our model with additional modifications to the training of the encoder for the reasons explained below.

There is a drawback to incorporating attribute information in the way described above \citep{makhzani2015adversarial} when the purpose of the model is to edit specific attributes, rather than to synthesize samples from a particular category. We observe that in this \textit{naive} implementation of conditional VAEs, varying the attribute (or label) vector, $\hat{y}$, for a fixed $\hat{z}$ can result in unpredictable changes in synthesized data samples, $\hat{x}$. Consider for example the case where, for a fixed $\hat{z}$, modifying $\hat{y}$ does not result in any change in the intended  corresponding attribute. This suggests that information about the attribute one wishes to edit, $y$, is partially contained in $\hat{z}$ rather than solely in $\hat{y}$. Similar problems have been discussed and addressed to some extent in the GAN literature \citep{chen2016infogan, mirza2014conditional,odena2016conditional}, where it has been observed that label information in $\hat{y}$ is often ignored during sample synthesis. 

In general, one may think that $\hat{z}$ and $\hat{y}$ should be independent. However, if attributes, $y$, that should be described by $\hat{y}$ remain unchanged for a reconstruction where only $\hat{y}$ is changed, this suggests that $\hat{z}$ contains most of the information that should have been encoded within $\hat{y}$. We propose a process to separate the information about $y$ from $\hat{z}$ using a mini-max optimization involving $y$, $\hat{z}$, the encoder $E_\phi$, and an auxiliary network $A_\psi$. We refer to our proposed process as `Adversarial Information Factorization'.

\subsection{Adversarial Information Factorization}

For a given image of a face, $x$, we would like to describe the face using a latent vector, $\hat{z}$, that captures the identity of the person, along with a single unit vector, $\hat{y}$, that captures the presence, or absence, of a single desired attribute, $y$. If a latent encoding, $\hat{z}$, contains information about the desired attribute, $y$, that should instead be encoded within the attribute vector, $\hat{y}$, then a classifier should be able to accurately predict $y$ from $\hat{z}$. Ideally, $\hat{z}$ contains no information about $y$ and so, ideally, a classifier should not be able to predict $y$ from $\hat{z}$. We propose to train an auxiliary network to predict $y$ from $\hat{z}$ accurately while updating the encoder of the VAE to output $\hat{z}$ values that cause the auxiliary network to fail. If $\hat{z}$ contains no information about the desired attribute, $y$, that we wish to edit, then the information can instead be conveyed in $\hat{y}$ since $\hat{x}$ must still contain that information in order to minimize reconstruction loss. We now formalize these ideas.

\section{Method}

In what follows, we explain our novel approach to training the encoder of a VAE, to factor (separate) out information about $y$ from $\hat{z}$, such that $H(y|\hat{z}) \approx H(y)$. We integrate this novel factorisation method into a VAE-GAN. The GAN component of the model is incorporated only to improve image quality. Our main contribution is our proposed adversarial method for factorising the label information, $y$, out of the latent encoding, $\hat{z}$.

\subsection{Model Architecture}
\label{sec:model_arch}
\begin{figure}[h!]
\centering
\begin{subfigure}{0.7\textwidth}
    \centering
    \includegraphics[width=\textwidth]{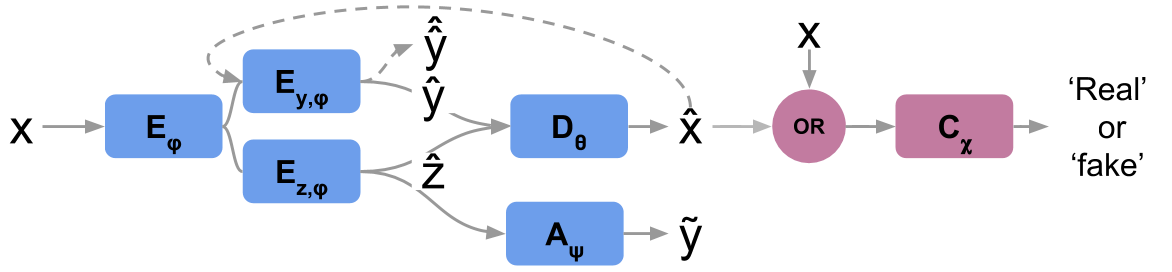}
    \caption{\textbf{Current work}}
\end{subfigure}
\begin{subfigure}{0.65\textwidth}
    \centering
    \includegraphics[width=\textwidth]{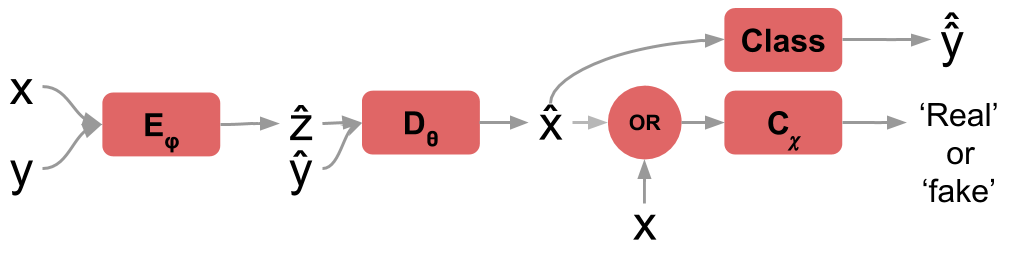}
    \caption{\textbf{Previous work} \citep{bao2017cvae}}
\end{subfigure}
    \caption{\textbf{(a) Current work (adversarial information factorization)} This figure shows our model where the core, shown in blue, is a VAE with information factorization. Note that $E_\phi$ is split in two, $E_{z,\phi}$ and $E_{y, \phi}$, to obtain both a latent encoding, $\hat{z}$, and the label, $\hat{y}$, respectively. $D_\theta$ is the decoder and $A_\psi$ the auxiliary network. The pink blocks show how a GAN architecture may be incorporated by placing a discriminator, $C_\chi$, after the encoder, $E_\phi$, and training $C_\chi$ to classify decoded samples as ``fake'' and samples from the dataset as ``real''. Finally, the dashed lines show how decoded samples may be passed back through the encoder to obtain a label, $\hat{\hat{y}}$, which may be used to obtain a gradient, that contains label information, for updating the decoder of the VAE. For simplicity, the $KL$ regularization is not shown in this figure. \textbf{(b) Previous work: cVAE-GAN} \citep{bao2017cvae} Architecture most similar to our own. Note that there is no auxiliary network performing information factorization and a label, $\hat{\hat{y}}$, is predicted only for the reconstructed image, rather than for the input image ($\hat{y}$).}
\label{fig:arch}
\end{figure}

A schematic of our architecture is presented in Figure \ref{fig:arch}. In addition to the encoder, $E_\phi$, decoder, $D_\theta$, and discriminator, $C_\chi$, we introduce an auxiliary network, $A_\psi : \hat{z} \rightarrow \tilde{y}$, whose purpose is described in detail in Section \ref{sec:IS}. We use $\hat{\hat{y}}$ to indicate the predicted label of a reconstructed data sample. Additionally, we incorporate a classification model into the encoder so that our model may easily be used to perform classification tasks.



The parameters of the decoder, $\theta$, are updated with gradients from the following loss function:
\begin{equation}
\mathcal{L}_{dec} = \mathcal{L}_{rec} + \beta \hat{\mathcal{L}}_{class} - \delta \mathcal{L}_{gan}
\end{equation}

where $\beta$ and $\delta$ are regularization coefficients, $\mathcal{L}_{rec} = L_{bce}(\hat{x}, x)$ is a reconstruction loss and $\hat{\mathcal{L}}_{class} = L_{bce}(\hat{\hat{y}}, y)$ is a classification loss on reconstructed data samples. The classification loss, $\hat{\mathcal{L}}_{class}$, provides a gradient containing label information to the decoder, which otherwise the decoder would not have \citep{chen2016infogan}. The GAN loss is given by $\mathcal{L}_{gan} = \frac{1}{3} [ L_{bce}(y_{real}, C_\chi(x)) + L_{bce}(y_{fake}, C_\chi(D_\theta(E_\phi(x)))) + L_{bce}(y_{fake}, C_\chi(D_\theta(z)))]$, where $y_{real}$ and $y_{fake}$ are vectors of ones and zeros respectively. Note that $L_{bce}$ is the binary cross-entropy loss given by $L_{bce}(a,b) = \frac{1}{M} \sum_{i=1}^M a \log b + (1 - a) \log (1 - b)$. The discriminator parameters, $\chi$, are updated to minimize $\mathcal{L}_{gan}$.




The parameters of the encoder, $\phi$, intended for use in synthesizing images from a desired \textbf{category}, may be updated by minimizing the following function:

\begin{equation} \label{eqn:enc-noIF}
\mathcal{L}_{enc} = \mathcal{L}_{rec}  + \alpha \mathcal{L}_{KL} + \beta \hat{\mathcal{L}}_{class} + \rho \mathcal{L}_{class} - \delta \mathcal{L}_{gan}
\end{equation}

where $\alpha$ and $\rho$ are additional regularization coefficients; $\mathcal{L}_{KL} = KL[q_\phi(z|x) || p(z)]$ and $\mathcal{L}_{class} = L_{bce}(\hat{y}, y)$ is the classification loss on the input image. Unfortunately, the loss function in Equation (\ref{eqn:enc-noIF}) is not sufficient for training an encoder used for attribute manipulation. For this, we propose an additional network and cost function, as described below.

\subsection{Adversarial Information Factorisation}
\label{sec:IS}

To factor label information, $y$, out of $\hat{z}$ we introduce an additional auxiliary network, $A_\psi$, that is trained to correctly predict $y$ from $\hat{z}$. The encoder, $E_\phi$, is simultaneously updated to promote $A_\psi$ to make incorrect classifications. In this way, the encoder is encouraged {\em not} to place attribute information, $y$, in $\hat{z}$. This may be described by the following mini-max objective:

\begin{equation} \label{eqn:gan}
\begin{split}
\min_\psi \max_\phi L_{bce}( A_\psi(E_{z,\phi}(x)), y) = \min_\psi \max_\phi L_{bce}(\tilde{y}, y)  = \min_\psi \max_\phi \mathcal{L}_{aux}
\end{split}
\end{equation}
where $E_{z,\phi(x)}$ is the latent output of the encoder.

Training is complete when the auxiliary network, $A_\psi$, is maximally confused and cannot predict $y$ from $\hat{z} = E_{z,\phi}(x)$, where $y$ is the true label of $x$. The encoder loss is therefore given by:
\begin{equation}
\mathcal{L}_{IFcVAE-GAN} = \mathcal{L}_{enc} - \mathcal{L}_{aux}
\end{equation}

We call the conditional VAE-GAN trained in this way an Information Factorization cVAE-GAN (IFcVAE-GAN). The training procedure is presented in Algorithm \ref{alg:train}.

\begin{algorithm}
\caption{\textbf{Training Information Factorization cVAE-GAN (IFcVAE-GAN)}: The prior, $p(z)=\mathcal{N}(\mathbf{0},I)$.}
\begin{algorithmic}[1]
\Procedure{Training cVAE with information factorization}{}
\For{$i$ in $range(N)$}  \Comment{$N$ is no. of epochs}
\State \Comment{forward pass through all networks}
\State $x \sim \mathcal{D}$ \Comment {$\mathcal{D}$ is the training data}
\State $z \sim p(z)$
\State $\hat{z}, \hat{y} \gets E_\phi(x)$ 
\State $\hat{x} \gets  D_\theta(\hat{y},\hat{z})$ 
\State $\tilde{y} \gets A_\psi(\hat{z})$ \Comment{output of the auxiliary network}
\State \# Calculate updates, $u$
\State \# do updates
\State $\theta \gets \text{RMSprop}(\theta, -\nabla_\theta \mathcal{L}_{dec})$
\State $\phi \gets \text{RMSprop}(\phi, -\nabla_\phi \mathcal{L}_{IFcVAE-GAN})$   \Comment{update the encoder to max. $\mathcal{L}_{aux}$}
\State $\chi \gets \text{RMSprop}(\chi, +\nabla_\chi \mathcal{L}_{gan})$
\State $\psi \gets \text{RMSprop}(\psi, -\nabla_\psi \mathcal{L}_{aux})$ \Comment{update auxiliary network to min. $\mathcal{L}_{aux}$}

\EndFor
\EndProcedure
\end{algorithmic}
\label{alg:train}
\end{algorithm}

\subsection{Attribute Manipulation} \label{sec:attribute_manipulation}
To edit an image such that it has a desired attribute, we encode the image to obtain a $\hat{z}$, the identity representation, append it to our desired attribute label, $\hat{y} \gets y$, and pass this through the decoder. We use $\hat{y}=0$ and $\hat{y}=1$ to synthesize samples in each \textit{mode} of the desired attribute e.g. `Smiling' and `Not Smiling'. Thus, attribute manipulation becomes a simple `switch flipping' operation in the representation space.

\section{Results} 
In this section, we show both quantitative and qualitative results to evaluate our proposed model. We begin by quantitatively assessing the contribution of each component of our model in an ablation study. Following this we perform facial attribute classification using our model. We use a standard deep convolutional GAN, DCGAN, architecture for the ablation study \citep{radford2015unsupervised}, and subsequently incorporate residual layers \citep{he2016deep} into our model in order to achieve competitive classification results compared with a state of the art model \citep{zhuang2018multi}. We finish with a qualitative evaluation of our model, demonstrating how our model may be used for image attribute editing. For our qualitative results we continue to use the same residual networks as those used for classification, since these also improved visual quality.

We refer to any cVAE-GAN that is trained without an $\mathcal{L}_{aux}$ term in the cost function as a naive cVAE-GAN and a cVAE-GAN trained with the $\mathcal{L}_{aux}$ term as an Information Factorization cVAE-GAN (IFcVAE-GAN).



\subsection{Quantifying contributions of each component to the final model}
\label{sec:quant_results}



Table \ref{table:contributions} shows the contribution of each component of our proposed model. We consider reconstruction error and classification accuracy on synthesized data samples. Smaller reconstruction error indicates better reconstruction, and larger classification values ($\hat{\mathcal{C}}_{Smiling}$ and $\hat{\mathcal{C}}_{Not-Smiling}$) suggest better control over attribute changes. To obtain $\hat{\mathcal{C}}_{Smiling}$ and $\hat{\mathcal{C}}_{Not-Smiling}$ values, we use an \textbf{independent} classifier, trained on real data samples to classify `Smiling' $(y=1)$ vs. `Not Smiling' $(y=0)$. We apply the trained classifier to two sets of image samples synthesized using $\hat{y}=0$ and $\hat{y}=1$. If the desired attributes are changed, the classification scores should be high for both sets of samples. Whereas if the desired attributes remain unchanged, the classifier is likely to perform well on only one of the sets, indicating that the attribute was not edited but fixed. Note that all original test data samples for this experiment were from the `Smiling' category. The results are shown in Table \ref{table:contributions}, where the classification scores ($\hat{\mathcal{C}}_{Smiling}$, $\hat{\mathcal{C}}_{Not Smiling}$) may be interpreted as the proportion of samples with the desired attributes and the MSE error interpreted as the fidelity of reconstruction. From Table \ref{table:contributions}, we make the following observations:

\textbf{Effect of $\hat{\mathcal{L}}_{class}$:} Using  $\hat{\mathcal{L}}_{class}$ does not provide any clear benefit. We explored the effect of including this term since a similar approach had been proposed in the GAN literature \citep{chen2016infogan,odena2016conditional} for conditional image synthesis (rather than attribute editing). To the best of our knowledge, this approach has not been used in the VAE literature. This term is intended to maximise $I(x,y)$ by providing a gradient containing label information to the decoder, however, it does not contribute to the factorization of attribute information, $y$, from $\hat{z}$.

\textbf{Effect of Information Factorization:}  Without our proposed  $\mathcal{L}_{aux}$ term in the encoder loss function, the model fails completely to perform attribute editing. Since $\hat{\mathcal{C}}_{Smiling}$ + $\hat{\mathcal{C}}_{Not-Smiling}$ $\approx 100\%$, this strongly suggests that samples are synthesized independently of $\hat{y}$ and that the synthesized images are the same for $\hat{y}=0$ and $\hat{y}=1$.

\textbf{Effect of $\hat{\mathcal{L}}_{class}$ on its own:} For completeness, we also evaluated our model without $\mathcal{L}_{aux}$ but with $\hat{\mathcal{L}}_{class}$ to test the effect of $\hat{\mathcal{L}}_{class}$ on its own. Though similar approaches have been successful for category conditional image synthesis, it was not as successful on the attribute editing task. Similarly, as above, $\hat{\mathcal{C}}_{Not-Smiling}$ + $\hat{\mathcal{C}}_{Smiling}$ $\approx 100\%$, suggesting that samples are synthesized independently of $\hat{y}$. Furthermore, $\hat{\mathcal{C}}_{Not-Smiling}$ $=0.0$, which suggests that none of the synthesized images had the desired attribute $\hat{y}=0$ (`Not Smiling'), i.e. all samples are with the attribute `Smiling'. This supports the use of $\mathcal{L}_{aux}$, when training models for attribute editing, over $\hat{\mathcal{L}}_{class}$ despite the promotion of the latter in the GAN literature \citep{chen2016infogan,odena2016conditional} for category specific sample synthesis.

\begin{table}[h!]
\centering
\caption{\textbf{What are the essential parts of the IFcVAE-GAN?} This table shows how novel components of the IFcVAE-GAN loss function affect mean squared (reconstruction) error, MSE, and the ability to edit facial attributes in an image. Ability to edit attributes is quantified by a pair of classification accuracies, $\hat{\mathcal{C}}_{Smiling}$ and $\hat{\mathcal{C}}_{Not-Smiling}$, on samples synthesized with $\hat{y}=1$ and $\hat{y}=0$ respectively. These values may be thought of as the proportion of synthesized images that have the desired attribute. We used hyper-parameters: $\{\rho=0.1, \delta=0.1, \alpha=0.2, \beta=0.0\}$.}
\begin{tabular}{r l l l}
\toprule

Model & MSE & $\hat{\mathcal{C}}_{Not-Smiling}$ & $\hat{\mathcal{C}}_{Smiling}$ \\
\midrule
Ours (without $\hat{\mathcal{L}}_{class}$) & 0.028 & 81.3\% & 100.0\%\\ 
With $\hat{\mathcal{L}}_{class}$& 0.028 & 93.8\% & 93.8\% \\
Without $\mathcal{L}_{aux}$, without $\hat{\mathcal{L}}_{class}$ &  0.028 & 18.8\% & 81.3\% \\
Without $\mathcal{L}_{aux}$, with $\hat{\mathcal{L}}_{class}$&  0.027 & 0.0\% & 100.0\% \\ 
\bottomrule
\end{tabular}
\label{table:contributions}
\end{table}

\subsection{Facial Attribute Classification} \label{sec:class}

We have proposed a model that learns a representation, $\{\hat{z}, \hat{y}\}$, for faces such that the identity of the person, encoded in $\hat{z}$, is factored from a particular facial attribute. We achieve this by minimizing the mutual information between the identity encoding and the facial attribute encoding to ensure that $H(y|\hat{z}) \approx H(y)$, while also training $E_{y, \phi}$ as an attribute classifier. Our training procedure encourages the model to put all label information into $\hat{y}$, rather than $\hat{z}$. This suggests that our model may be useful for facial attribute classification.

\begin{figure}[h!]
\centering
\includegraphics[width=\textwidth]{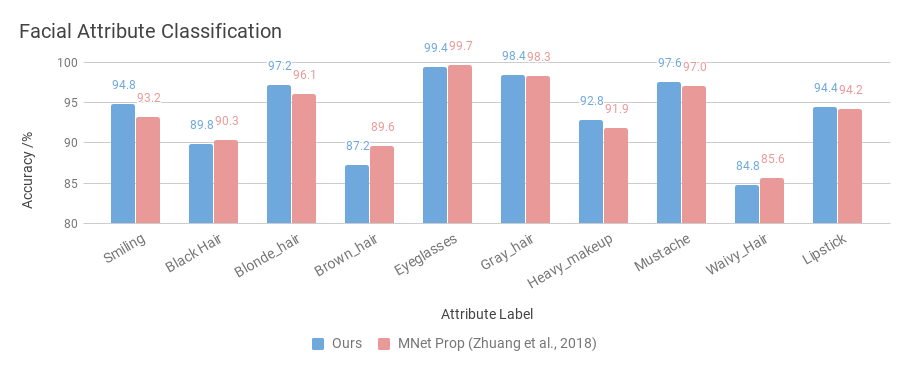} 
\caption{\textbf{Facial Attribute Classification.} We compare the performance of our classifier, $E_{y,\phi}$, to a state of art classifier \citep{zhuang2018multi}. Our model outperforms that of \cite{zhuang2018multi} for $6$ out of $10$ attributes and achieves comparable results for most other attributes.}
\label{fig:state_of_art}
\end{figure}

To further illustrate that our model is able to separate the representation of particular attributes from the representation of the person's identity, we can measure the model's ability, specifically the encoder, to classify facial attributes. We proceed to use $E_{y, \phi}$ directly for facial attribute classification and compare the performance of our model to that of a state of the art classifier proposed by \cite{zhuang2018multi}. Results in Figure \ref{fig:state_of_art} show that our model is highly competitive with a state of the art facial attribute classifier, outperforming \cite{zhuang2018multi} on $6$ out of $10$ categories and remaining competitive in most other attributes. These results demonstrate that the model is effectively factorizing out information about the attribute from the identity representation.




\subsection{Qualitative Results}

\label{sec:qual_results}

In this section, we focus on attribute manipulation (described previously in Section \ref{sec:attribute_manipulation}). Briefly, this involves reconstructing an input image, $x$, for different attribute values, $\hat{y} \in \{0,1\}$.

\begin{figure}[h!]
\centering
\begin{subfigure}{0.7\textwidth}
\centering
\includegraphics[width=0.7\textwidth]{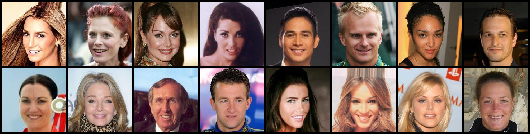} 
\caption{Original Smiling Faces.}
\end{subfigure}
\begin{subfigure}{0.49\textwidth}
\centering
\includegraphics[width=\textwidth]{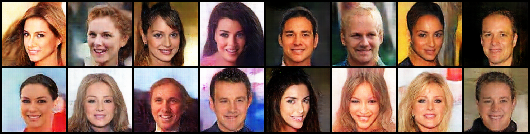} 
\caption{Smiling (ours).}
\end{subfigure}
\begin{subfigure}{0.49\textwidth}
\centering
\includegraphics[width=\textwidth]{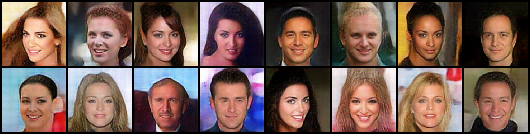} 
\caption{Smiling \citep{bao2017cvae}.}
\end{subfigure}
\begin{subfigure}{0.49\textwidth}
\centering
\includegraphics[width=\linewidth]{graphics/rec_0_Ex191} 
\caption{Not Smiling (ours).}
\end{subfigure}
\begin{subfigure}{0.49\textwidth}
\centering
\includegraphics[width=\linewidth]{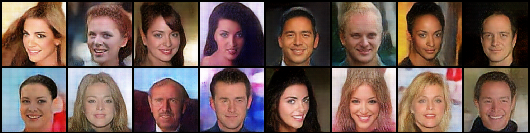} 
\caption{(Failed) Not Smiling \citep{bao2017cvae}.}
\end{subfigure}
\caption{\textbf{Reconstructions, `Smiling' and `Not Smiling'.} The goal here was to reconstruct the face, changing only the desired `Smiling' attribute. This demonstrates how other conditional models \citep{bao2017cvae} may fail at the image attribute editing task, when high quality reconstructions are required. Both models are trained with the same optimizers and hyper-parameters.}
\label{fig:bestA}
\end{figure}

We begin by demonstrating how a naive cVAE-GAN \citep{bao2017cvae} may fail to edit desired attributes, particularly when it is trained to achieve low reconstruction error. The work of \cite{bao2017cvae} focused solely on the ability to synthesise images with a desired attribute, rather than to reconstruct a particular image and specifically edit one of its attributes. It is challenging to learn a representation that both preserves identity and allows factorisation \citep{higgins2016beta}. Figure \ref{fig:bestA}(c,e) shows reconstructions when setting $\hat{y}=0$ for `Not Smiling' and $\hat{y}=1$ for `Smiling'. We found that the naive cVAE-GAN \citep{bao2017cvae} failed to synthesise samples with the desired target attribute `Not Smiling'. This failure demonstrates the need for models that can deal with both reconstruction and attribute-editing. Note that we achieve good reconstruction by reducing weightings on the $KL$ and GAN loss terms, using $\alpha=0.005$ and $\delta=0.005$ respectively. We trained the model using RMSProp \citep{tieleman2012lecture} with momentum $=0.5$ in the discriminator.

We train our proposed IFcVAE-GAN model using the same optimiser and hyper-parameters that were used for the \cite{bao2017cvae} model above. We also used the same number of layers (and residual layers) in our encoder, decoder and discriminator networks as those used by \cite{bao2017cvae}. Under this set-up, we used the following additional hyper-parameters: $\{\beta=0.0, \rho=1.0\}$ in our model. Figure \ref{fig:bestA} shows reconstructions when setting $\hat{y}=0$ for `Not Smiling' and $\hat{y}=1$ for `Smiling'. In contrast to the naive cVAE-GAN \citep{bao2017cvae}, our model is able to achieve good reconstruction, capturing the identity of the person, while also being able to change the desired attribute. Table \ref{table:comparison} shows that the model was able to synthesize images with the `Not Smiling' attribute with a $98\%$ success rate, compared with a $22\%$ success rate using the naive cVAE-GAN \cite{bao2017cvae}.

\begin{table}[h!]
\centering
\caption{\textbf{Comparing our model, the IFcVAE-GAN, to the naive cVAE-GAN \cite{bao2017cvae}}. Ability to edit attributes is quantified by a pair of classification accuracies, $\hat{\mathcal{C}}_{Smiling}$ and $\hat{\mathcal{C}}_{Not-Smiling}$, on samples synthesized with $\hat{y}=1$ and $\hat{y}=0$ respectively. These values may be thought of as the proportion of synthesized images that have the desired attribute. We see that both models achieve comparable (MSE) reconstruction errors, however, only our model is able to synthesize images of faces without smiles. A complete ablation study for this model (with residual layers) is given in the appendix (Table \ref{table:contributions_resNets}).}.
\begin{tabular}{r l l l}
\toprule

Model & MSE & $\hat{\mathcal{C}}_{Not-Smiling}$ & $\hat{\mathcal{C}}_{Smiling}$ \\
\midrule
Ours (with residual layers) & 0.011 & 98\% & 100\%\\
\cite{bao2017cvae} (with residual layers) & 0.011  & 22\% & 85\%\\ 
\bottomrule
\end{tabular}
\label{table:comparison}
\end{table}


\subsection{Editing Other Facial Attributes}
\begin{figure}[h!]
\centering
\includegraphics[width=\textwidth]{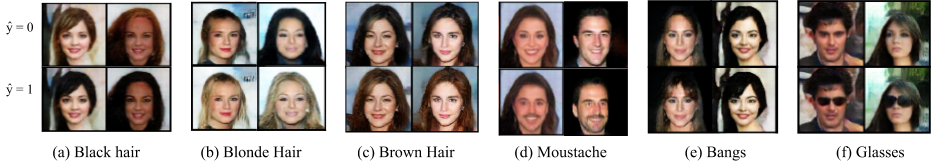} 
\caption{\textbf{Editing other attributes.} We obtain a $\hat{z}$, the identity representation, by passing an image, $x$ through the encoder. We append $\hat{z}$ with a desired attribute label, $\hat{y} \gets y$, and pass this through the decoder. We use $\hat{y}=0$ and $\hat{y}=1$ to synthesize samples in each \textit{mode} of the desired attribute}\label{fig:other}
\end{figure}

In this section we apply our proposed method to manipulate other facial attributes where the initial samples, from which the $\hat{z}$'s are obtained, are test samples whose labels are $y=1$ indicating the presence of the desired attribute (e.g. `Blonde Hair'). In Figure \ref{fig:other}, we observe that our model is able to both achieve high quality reconstruction and edit the desired attributes.

We have presented the novel IFcVAE-GAN model, and (1) demonstrated that our model learns to factor attributes from identity, (2) performed an ablation study to highlight the benefits of using an auxiliary classifier to factorize the representation and (3) shown that our model may be used to achieve competitive scores on a facial attribute classification task. We now discuss this work in the context of other related approaches.

\section{Comparison to Related Work} \label{sec:related_work}

We have used adversarial training (involving an auxiliary classifier) to factor attribute label information, $y$, out of the encoded latent representation, $\hat{z}$. Schmidhuber \citep{schmidhuber2008learning} performs similar factorization of the latent space, ensuring that each component of the encoding is independent. This is achieved by learning an encoding such that other elements in the encoding may not be predicted from a subset of remaining elements. We use related concepts, with additional class label information, and incorporate the encoding in a generative model.

Our work has the closest resemblance to the cVAE-GAN  architecture (see Figure \ref{fig:arch}) proposed by \cite{bao2017cvae}. cVAE-GAN is designed for synthesizing samples of a particular class, rather than manipulating a single attribute of an image from a class. In short, their objective is to synthesize a ``Hathway'' face, whereas our objective would be to make ``Hathway smiling'' or ``Hathway not smiling'', which has different demands on the type of factorization in the latent representation. Separating categories is a simpler problem since it is possible to have distinct categories and changing categories may result in more noticeable changes in the image. Changing an attribute requires a specific and targeted change with minimal changes to the rest of the image. Additionally, our model simultaneously learns a classifier for input images unlike the work by \cite{bao2017cvae}.

In a similar vein to our work, \cite{antipov2017face} acknowledge the need for ``identity preservation'' in the latent space. They achieve this by introducing an identity classification loss between an input data sample and a reconstructed data sample, rather than trying to separate information in the encoding itself. Similar to our work, \cite{larsen2015autoencoding} use a VAE-GAN architecture. However, they do not condition on label information and their image ``editing'' process is not done in an end-to-end fashion \footnote{\cite{larsen2015autoencoding} traverse the latent space along an attribute vector found by taking the mean difference between encodings of several samples with the same attribute. Additionally, in Figure 5 of \cite{larsen2015autoencoding}, changing one attribute results in other attributes changing, for example in the bottom row when changing the `blonde hair' attribute, the woman's make-up changes too.}.

Our work highlights an important difference between category conditional image synthesis \citep{bao2017cvae} and attribute editing in images: what works for category conditional image synthesis may not work for attribute editing. Furthermore, we have shown (Section \ref{sec:quant_results}) that for attribute editing to be successful, it is necessary to factor label information out of the latent encoding.

In this paper, we have focused on latent space generative models, where a small change in latent space results in a semantically meaningful change in image space. Our approach is orthogonal to a class of image \textit{editing} models, called ``image-to-image'' models, which aim to learn a single latent representation for images in different domains. Recently, there has been progress in image-to-image domain adaptation, whereby an image is translated from one domain (e.g. a photograph of a scene) to another domain (e.g. a painting of a similar scene) \citep{zhu2017unpaired, liu2017unsupervised, liu2016coupled}. Image-to-image methods may be used to translate smiling faces to non-smiling faces \citep{liu2017unsupervised, liu2016coupled}, however, these models \citep{liu2017unsupervised, liu2016coupled} require significantly more resources than ours\footnote{While our approach requires a single generative model, the approaches of \cite{liu2017unsupervised, liu2016coupled} require a pair of generator networks, one for each domain.}. By performing factorization in the latent space, we are able to use a single generative model, to edit an attribute by simply changing a single unit of the encoding, $y$, from $0$ to $1$ or vice versa.

\section{Conclusion}
We have proposed a novel perspective and approach to learning representations of images which subsequently allows elements, or attributes, of the image to be modified. We have demonstrated our approach on images of the human face, however, the method is generalisable to other objects. We modelled a human face in two parts, with a continuous latent vector that captures the identity of a person and a binary unit vector that captures a facial attribute, such as whether or not a person is smiling. By modelling an image with two separate representations, one for the object and the other for the object's attribute, we are able to change attributes without affecting the identity of the object. To learn this factored representation we have proposed a novel model aptly named Information Factorization conditional VAE-GAN. The model encourages the attribute information to be factored out of the identity representation via an adversarial learning process. Crucially, the representation learned by our model \textbf{both} captures identity faithfully and facilitates accurate and easy attribute editing without affecting identity. We have demonstrated that our model performs better than pre-existing models intended for category conditional image synthesis (Section \ref{sec:qual_results}), and have performed a detailed ablation study (Table \ref{table:contributions}) which confirms the importance and relevance of our proposed method. Indeed, our model is highly effective as a classifier, achieving state of the art accuracy on facial attribute classification for several attributes (Figure \ref{fig:state_of_art}). Our approach to learning factored representations for images is both a novel and important contribution to the general field of representation learning.




\bibliography{iclr2019_conference}
\bibliographystyle{iclr2019_conference}

\appendix
\section*{Appendix}





\subsection*{Ablation Study For Our Model With Residual Layers}

\begin{figure}[h]
\centering
\begin{subfigure}{0.7\textwidth}
\centering
\includegraphics[width=0.7\textwidth]{graphics/original.png} 
\caption{Original Smiling Faces.}
\end{subfigure}
\begin{subfigure}{0.49\textwidth}
\centering
\includegraphics[width=\textwidth]{graphics/rec_1_Ex191.png} 
\caption{Smiling (ours).}
\end{subfigure}
\begin{subfigure}{0.49\textwidth}
\centering
\includegraphics[width=\textwidth]{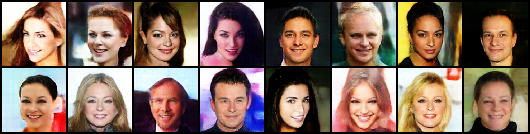} 
\caption{Smiling (without $\mathcal{L}_{aux}$).}
\end{subfigure}
\begin{subfigure}{0.49\textwidth}
\centering
\includegraphics[width=\linewidth]{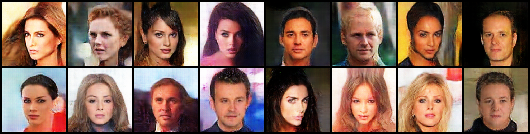} 
\caption{Not Smiling (ours).}
\end{subfigure}
\begin{subfigure}{0.49\textwidth}
\centering
\includegraphics[width=\linewidth]{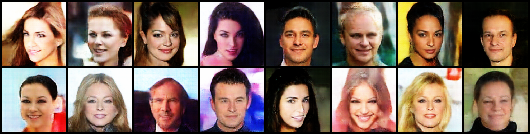} 
\caption{(Failed) Not Smiling  (without $\mathcal{L}_{aux}$).}
\end{subfigure}
\caption{\textbf{Reconstructions, `Smiling' and `Not Smiling', with and without $\mathcal{L}_{aux}$.} The goal here was to reconstruct the face, changing only the desired `Smiling' attribute. This figure demonstrates the need for the $\mathcal{L}_{aux}$ term in our model. Both models are trained with the same optimizers and hyper-parameters.}
\label{fig:bestB}
\end{figure}

For completeness we include a table (Table \ref{table:contributions_resNets}) demonstrating an ablation study for our model with the residual network architecture discussed in Section \ref{sec:qual_results}, note that this is the same architecture that was used by \cite{bao2017cvae}. Table \ref{table:contributions_resNets} and additionally, Figure \ref{fig:bestB}, demonstrate the need for the $L_{aux}$ loss and shows that increased regularisation reduces reconstruction quality. The table also shows that there is no significant benefit to using the $\hat{\mathcal{L}}_{class}$ loss. These findings are consistent with those of the ablation study in the main body of the text for the IFcVAE-GAN with a the GAN architecture of \cite{radford2015unsupervised}.

\begin{table}
\centering
\caption{\textbf{What are the essential parts of the IFcVAE-GAN (with residual layers)?} This table shows how novel components of the IFcVAE-GAN loss function affect mean squared (reconstruction) error, MSE, and the ability to edit facial attributes in an image. Ability to edit attributes is quantified by a pair of classification accuracies, $\hat{\mathcal{C}}_{Smiling}$ and $\hat{\mathcal{C}}_{Not-Smiling}$, on samples synthesized with $\hat{y}=1$ and $\hat{y}=0$ respectively. These values may be thought of as the proportion of synthesized images that have the desired attribute. We use hyper-parameters: $\{\rho=0.1, \delta=0.005, \alpha=0.005, \beta=0.0, momentum=0.5\}$. We also show classification accuracy (Acc.) of $E_{y,\phi}$. \newline
*\small{Note that the model of \cite{bao2017cvae} does not incorporate a classifier}.}
\begin{tabular}{r l l l l }
\toprule

Model & MSE & $\hat{\mathcal{C}}_{Not-Smiling}$ & $\hat{\mathcal{C}}_{Smiling}$ & Acc. ($E_{y,\phi}$)\\
\midrule
Ours (with residual layers) $(\alpha=0.005)$ & 0.011 & 98\% & 100.0\% & 92\%\\   
Higher levels of regularization $(\alpha=0.1)$  & 0.020 & 100\% & 100\% & 92\%\\
With $\hat{\mathcal{L}}_{class}$, $(\alpha=0.005)$& 0.010 & 96\% & 100\% & 91\%\\
Without $\mathcal{L}_{aux}$, $(\alpha=0.005)$ &  0.013 & 28\% & 91\% & 91\%\\
Without $\mathcal{L}_{aux}$, with $\hat{\mathcal{L}}_{class}$, $(\alpha=0.005)$,  & 0.019 & 33\% & 96\% & 89\% \\
\cite{bao2017cvae}, $(\alpha=0.005)$ & 0.011 & 22\% & 85\% & n/a* \\
\bottomrule
\end{tabular}
\label{table:contributions_resNets}
\end{table}

\end{document}